\pdfoutput=1
\documentclass[sigconf]{aamas} 

\usepackage{todonotes}

\usepackage{flushend}
\setcopyright{ifaamas}  
\acmDOI{doi}  
\acmISBN{}  
\acmConference[AAMAS'22]{Proc.\@ of the 21st International Conference
on Autonomous Agents and Multiagent Systems (AAMAS 2022),
P.~Faliszewski, V.~Mascardi, C.~Pelachaud,
M.E.~Taylor (eds.)}{May 2022}{Auckland, New Zealand} 
\acmYear{2022}  
\copyrightyear{2022}  
\acmPrice{}
\acmISBN{}

\title{Multiagent Model-based Credit Assignment for \\Continuous Control}

\author{
Dongge Han$^{1}$,
Chris Xiaoxuan Lu$^{2}$,
Tomasz Michalak$^{3}$,
Michael	Wooldridge$^{1}$
}

\affiliation{
  \institution{$^{1}$ University of Oxford, Oxford, United Kingdom}
  \institution{$^{2}$ University at Edinburgh, Edinburgh, United Kingdom}
  \institution{$^{3}$ University at Warsaw, Warsaw, Poland}
  \country{}
}

\begin{abstract}
Deep reinforcement learning (RL) has recently shown great promise in robotic continuous control tasks. Nevertheless, prior research in this vein center around the centralized learning setting that largely relies on the communication availability among all the components of a robot. However, agents in the real world often operate in a decentralised fashion without communication due to latency requirements, limited power budgets and safety concerns. By formulating robotic components as a system of decentralised agents, this work presents a decentralised multiagent reinforcement learning framework for continuous control. To this end, we first develop a cooperative multiagent PPO framework that allows for centralized optimisation during training and decentralised operation during execution. However, the system only receives a global reward signal which is not attributed towards each agent. To address this challenge, we further propose a generic game-theoretic credit assignment framework which computes agent-specific reward signals. Last but not least, we also incorporate a model-based RL module into our credit assignment framework, which leads to significant improvement in sample efficiency. We demonstrate the effectiveness of our framework on experimental results on Mujoco locomotion control tasks. For a demo video please visit: \url{https://youtu.be/gFyVPm4svEY}.
\end{abstract}


\newcommand{\BibTeX}{\rm B\kern-.05em{\sc i\kern-.025em b}\kern-.08em\TeX}
\usepackage{xspace}
\usepackage{tabularx}
\usepackage{booktabs}

\newcommand{\mdp}{\mathcal{M}}

\newcommand{\states}{\ensuremath{\mathcal{S}}}
\newcommand{\actions}{\ensuremath{\mathcal{A}}}
\newcommand{\dynamics}{\ensuremath{P}}
\newcommand{\data}{\ensuremath{\mathcal{D}}}
\newcommand{\reward}{\ensuremath{r}}
\newcommand{\policy}{\mathbf{\pi}}

\newcommand{\tmodel}{\ensuremath{f_s}}
\newcommand{\rmodel}{\ensuremath{f_r}}
\newcommand{\mparam}{\ensuremath{\phi}}
\newcommand{\aparam}{\ensuremath{\theta}}
\newcommand{\vparam}{\ensuremath{\omega}}

\newcommand{\coalition}{\ensuremath{\mathcal{C}}}

\newcommand{\mc}{\ensuremath{\mathcal{MC}}}
\newcommand{\cv}{\ensuremath{v}}
\newcommand{\val}{\ensuremath{\psi}} 
\newcommand{\vshapley}{\ensuremath{\psi_{\textnormal{Shapley}}}}
\newcommand{\vbanzhaf}{\ensuremath{\psi_{\textnormal{Banzhaf}}}}

\newcommand{\weight}[2]{\ensuremath{w_{#1}}}

\newcommand{\mbshapley}{\texttt{MB-Shapley}\xspace}
\newcommand{\mbbanzhaf}{\texttt{MB-Banzhaf}\xspace}
\newcommand{\mbloo}{\texttt{MB-Loo}\xspace}
\newcommand{\qshapley}{\texttt{Q-Shapley}\xspace}
\newcommand{\mappo}{\texttt{MAPPO}\xspace}

\usepackage{bbm}
\usepackage[export]{adjustbox}

\usepackage{tikz}
\usetikzlibrary{fit}
\usetikzlibrary{arrows.meta}
\usepackage{amsmath,amsthm,mathtools,nccmath}
\usepackage{graphicx}
\usepackage{subfig}
\usepackage{pdfpages}
\usepackage{subfiles}
\usepackage{float}
\usepackage{natbib}
\usepackage{multirow}
\usepackage{graphicx}
\usepackage{xcolor}
\usepackage{algorithm}
\usepackage{algpseudocode}
\usepackage{wrapfig}
\usepackage{aligned-overset}
\usepackage{thm-restate}
\usepackage{enumitem}
\usepackage{booktabs}
\usepackage{pifont}

\begin{document}


\pagestyle{fancy}
\fancyhead{}


\maketitle 

\section{Introduction}
Reinforcement learning (RL) has recently shown many remarkable successes, e.g., in playing Atari and Go at a superhuman level~\cite{mnih2015human,silver2017mastering}. Recently, there have been wide research interest and significant advances in robotics learning using RL techniques~\cite{levine2016end,finn2016guided}. The topics of RL and classical control theory are closely related: both aim to find an optimal policy that optimizes an objective function, given a system represented by states and transition dynamics. Therefore,  RL algorithms have the potential of enabling robots to learn 
in complex real-world tasks such as locomotion, manipulation and navigation.
Unlike classical RL tasks, which have discrete action spaces and underlying state spaces (e.g. Atari and Go), problems in robotics often have high-dimensional continuous states and actions, and are often limited by real-world sample budgets~\cite{kober2013reinforcement}. To this end, prior research in robotic learning have developed RL algorithms capable of performing continuous control~\cite{lillicrap2015continuous, gu2017deep, schulman2017proximal, haarnoja2018soft}, and sample-efficient learning methods, e.g.,~\cite{andrychowicz2017hindsight,janner2019trust, feinberg2018model}.

Despite the success demonstrated by RL methods in continuous control, when applying RL algorithms to robotic applications, it is essential not to overlook real-life physical limits such as sensing noise and communication delays~\cite{kober2013reinforcement}. Specifically, prior RL approaches typically model a robot as a single, fully-centralised agent which observes the whole state consisting of observations from each body/joint and learns an optimal policy that outputs a combined action for all controllers. Though a centralized controller can compensate for the state of the whole system, when deployed in complex, unseen real-life settings, local observations may be noisy, and communications of the local observations between body components can be delayed due to physical limitations. Moreover, robots with a centralised controller do not scale up to large number of controllers and can easily become incapacitated if any sub-component is compromised. In such scenarios, a decentralised control system is desirable, where the robot controllers learn and execute their own policies. 
Fortunately, many real-life robots consist of multiple connected links which can be modelled as a multiagent system~\cite{wooldridge2009introduction}. 
To enable  coordination among agents with local observations and decentralised controllers, a standard approach is to perform centralized optimisation during training and decentralised operation during execution (CTDE)~\cite{foerster2018counterfactual, lowe2017multi}, i.e., the system can benefit from the state of the whole system during training, while acting in a decentralised manner. Using this framework, multiagent systems with RL agents have demonstrated coordination skills in complex coordination tasks such as Starcraft~\cite{samvelyan2019starcraft, rashid2018qmix, yu2021surprising}. 

Inspired by the above  framework, we apply multiagent RL to robotics which exhibits high-dimensional, continuous state and action spaces. Specifically, we extend Proximal Policy Optimisation (PPO)~\cite{schulman2017proximal}---an algorithm widely used in both discrete and continuous control tasks---to a fully cooperative multiagent framework with CTDE. 
However, as the robot interacts with the environment as a whole, \emph{only a global reward signal for the whole system is available while a reward function which signifies the contribution of each individual agent is absent.}

To address this problem, one can adopt solution concepts from cooperative game theory~\cite{chalkiadakis2011computational} that perform credit assignments for each agent, using its (marginal) contributions towards all possible coalitions of other agents. 
Several prior multiagent RL methods have used one such well-known solution concept, the Shapley value~\cite{chalkiadakis2011computational}), for credit assignment in applications with discrete action spaces (such as traffic junction~\cite{wang2020shapley}, Starcraft~\cite{li2021shapley} and cooperative sensing~\cite{xu2020learning}). However, \emph{no model of this kind has been proposed for continuous action spaces so far}. 
Our first contribution in this work is the formulation of a generic game-theoretic credit assignment framework for multiagent continuous control using semivalues~\cite{dubey1981value}---a wide family of solution concepts that encompass many common credit assignment methods such as the Shapley value, the Banzhaf value, etc. This formulation provides a useful tool for studying the underlying relationship and differences between the commonly used credit assignment methods in multiagent RL.

The formulation of game-theoretic credit assignment allows us to fairly evaluate the contribution of each agent. 
However, to compute the credit assignments, \emph{how can we obtain the value of different coalitions of agents, when only the value of the grand coalition is available (i.e., the global reward of the whole system)?} 
As we will discuss in Section~\ref{sub:mb}, if we simply apply the method to estimate values of coalitions that was proposed for models with discrete action spaces~\cite{li2021shapley} to continues ones, we would suffer from highly inaccurate results. This stems from a common problem in the multiagent robotics control domain, caused by the high dimensional continuous state and action spaces. To address this, we further propose a model-based framework for accurately estimating the values of coalitions of agents. This model-based framework greatly improves credit assignment in continuous control tasks and significantly boosts the rewards and sample-efficiency.

Finally, we empirically evaluate our proposed methods on MuJoCo robotic continuous control tasks. We demonstrate that our model-based game-theoretic credit assignment leads to significant improvements in rewards and sample-efficiency compared to a shared advantage function and model-free credit assignment methods.

\section{Background}

\noindent In this section, we introduce our notation and main definitions (see  Table~\ref{table:notations} for a summary of notation).

\subsection{Reinforcement Learning:}\label{subseq_background:rl}
A (single-agent) RL task is commonly defined by a reward function~\cite{sutton2018reinforcement}, and the goal of an agent is to learn an optimal policy which maximizes the cumulative future rewards. This can be modelled by the Markov decision process (MDP): $\mdp = \langle \states, \actions, \dynamics, \reward, \gamma \rangle$. At each time step $t$, an agent in state $s_t \in \states$, chooses an action $a_t \in \actions$, receives a reward $r_t = \reward(s_t,a_t)$, and transits to the next state $s_{t+1}\in \states$ according to an unknown transition dynamics $\dynamics:\states\times\actions\rightarrow\states$. The agent aims to maximise the future return $G_t = \sum_{k=0}^\infty{\gamma^k} r_{t+k+1}$ where $\gamma\in [0, 1]$ is a discount factor. We denote an agent's experience tuples as $\data_t = \langle s_t, a_t, s_{t+1}, r_t \rangle$. The policy of an agent is a mapping from states to probabilities of selecting a possible action, denoted as $\policy:\states\times\actions\rightarrow [0,1]$.

\subsubsection{Value Functions}\label{subsec:QV}
 The \emph{(state) value function}, denoted $V^\pi(s)$, is the expected return of an agent which starts in $s$ and follows policy $\pi$ thereafter:
\begin{equation}
    V^\pi(s) \coloneqq \mathbbm{E}_\pi[G_t|s_t=s] =  \mathbbm{E}_\pi[\sum_{k=0}^\infty \gamma^k r_{t+k+1}| s_t=s], \forall s\in \states.
\end{equation}
The \emph{(action) value function} $Q^\pi(s, a)$ defines the expected return when starting from $s$, taking action $a$, and following $\pi$ thereafter:
\begin{align}
    Q^\pi(s,a) &\coloneqq \mathbbm{E}_\pi[G_t|s_t=s, a_t=a]\nonumber\\ &\phantom{:}=  \mathbbm{E}_\pi[\sum_{k=0}^\infty \gamma^k r_{t+k+1}| s_t=s, a_t=a], \forall s\in \states.
\end{align}

\begin{table}
\caption{Notation}\label{table:notations}
\begin{tabularx}{\linewidth}{p{0.1\textwidth}X}
\toprule
    $i$, $t$, $k$ & Indices for agents, timesteps, iterations\\
    $\pi^i$ & Policy of agent $i$ (Sec.~\ref{subseq_background:rl},~\ref{sub:mappo})\\
    $G_t$ & Discounted return from $t$ onwards (Sec.~\ref{subseq_background:rl})\\
    $V^\pi(s), Q^\pi(s, \mathbf{a})$ & State and action-value functions (Sec.~\ref{subsec:QV})\\
    $A^\pi(s, \mathbf{a})$ & Advantage function (Sec.~\ref{subsec:PPO},~\ref{sub:mappo})\\
    $\vparam$, $\aparam$, $\mparam_s$, $\mparam_r$ & parameters of $V, Q, \tmodel, \rmodel$ (Sec.~\ref{sub:mappo},~\ref{sub:mb})\\
    $\cv^\coalition (s,\mathbf{a})$ & (Characteristic) value of coalition (Sec.~\ref{subsec:mc})\\
    $\Tilde{\mathbf{a}}_t$ & Actions $\mathbf{a}_t$ masked by coalitions (Sec.~\ref{subsec:mc})\\
    $\mc^i(\coalition, s, \mathbf{a})$ & Marginal contribution of agent $i$ to coalition $\coalition$ \\
    $\val^i(v)$, $p_c$ & semivalues and its probability indices (Sec.~\ref{subsec:semi})\\
    $\tmodel(s,\mathbf{a}),\rmodel(s,\mathbf{a})$ & dynamics and reward models (Sec.~\ref{subsec:model})\\
    $\hat{s}_{t+1}, \hat{r}_t$ & States/Rewards predicted by model (Sec.~\ref{subsec:model})\\
\bottomrule
\end{tabularx}
\end{table}

\subsubsection{Proximal Policy Optimisation (PPO)}\label{subsec:PPO}
PPO~\cite{schulman2017proximal} is an actor-critic policy gradient RL algorithm which is widely used in both discrete and continuous control tasks, due to its simplicity in implementation, sample complexity, and ease of hyperparameter tuning. Specifically, an \emph{actor} corresponds to the policy $\pi(a|s)$, while a \emph{critic} corresponds to the value function $V^\pi(s)$. The actor update is guided by an \emph{advantage function} $A(s, a)$ estimated using the critic. 
The advantage function $A^\pi(s,a)=Q^\pi(s,a)-V^\pi(s)$  measures
how much better (worse) taking action $a$ in state $s$ is compared with the policy’s default behavior. 
Intuitively, if the advantage $A(s, a)$ is positive, the actor parameters move towards the direction where this action becomes more likely. To improve stability, PPO avoids parameter updates that change the policy too much by clipping the ratio between old and new policies.
In practise, $A^\pi(s,a)$ can be estimated using methods such as general advantage estimation (GAE)~\cite{schulman2015high}, which makes use of value function $V^\pi(s)$ (the details of GAE can be found in~\cite{schulman2015high}).

\subsubsection{Centralised Training, Decentralised Execution}\label{subsec:CTDE} A standard practise in training a decentralised multiagent RL system is
Centralised Training, Decentralised Execution (CTDE)~\cite{foerster2018counterfactual}. 
For example, in a multiagent system with agents $N=\{1,\ldots, n\}$,  a \emph{centralised critic} (typically, action value functions $Q(\mathbf{s}_t, a_1, \ldots, a_n)$) 
and decentralised actors $(\pi_1, \ldots, \pi_n)$ are learned. 
The centralised critic takes as input the observations and joint actions of all agents, while the decentralised actors take the agents' local observations as input and outputs an action per agent. 
Compared with having decentralised critics (one per agent) which only conditions on the local observations and actions, the advantage of CTDE is that the centralised critic prevents an agent from experiencing un-stationary environment as a result of treating other learning agents as part of the environment. Moreover, the critic is only used during training; hence, the agents learn coordinated behaviours which can be executed in a decentralised manner during execution.


\noindent
\subsection{Cooperative Game Theory}\label{subsec:cgt}
A \emph{cooperative game} is given by a pair $G = (N, \cv)$, where $N = \{1,\ldots,n\}$ is the set of agents (players), and  $\cv\colon 2^N\rightarrow{\mathbb{R}}$ is the \emph{characteristic function}, which assigns a real value $\cv(\coalition)$ to every coalition $\coalition$ reflecting its performance.
We assume $\cv(\emptyset) = 0$.
Subsets of agents $\coalition \subseteq N$ are called \emph{coalitions}. 
To measure the contribution of an individual agent to a cooperative task, 
we use solution concepts~\cite{chalkiadakis2011computational} from cooperative game theory that evaluate the contribution of each agent $i \in N$ to the game. We will denote it $\val^i\in \mathbb{R}$. Specifically, we focus on a wide class of solution concepts called semivalues~\cite{dubey1981value}. For clarity, we will introduce the general definition of semivalues in  Section~\ref{sub:gt}. Before this, let us introduce the concept of \emph{marginal contribution} and two well-known semivalues: the Shapley and Banzhaf values. Intuitively, the marginal contribution of $i$ to coalition $\coalition$ is the difference that this agent makes towards $\coalition$ before and after joining it, i.e., $\mc^i(\coalition) \coloneqq \cv(\coalition\cup\{i\}) - \cv(\coalition)$.

The Shapley value~\cite{Shapley+2016+307+318} is the most common solution concept. It is defined as follows: 
        \begin{equation*}
        \vshapley^i(N, \cv) = 
        \sum_{S\subseteq N \setminus \{i\}}\frac{|\coalition|!(|N|-|\coalition| -1)!}{|N|!} \mc^i(\coalition).
        \end{equation*}
The Banzhaf value~\cite{banzhaf1964weighted} is another common solution concept: 
        \begin{equation*}
          \vbanzhaf^i(N, \cv) = \frac{1}{2^{|N|-1}}\sum_{\coalition\subseteq N \setminus \{i\}} \mc^i(\coalition).
        \end{equation*}
Intuitively, the Banzhaf value of $i$ is simply the average marginal contribution of this agent across all possible coalitions. The Shapley value is the weighted average, where the weight of each marginal contribution depends on the size of the coalition and the total number of agents.

\subsection{Kinematics Tree of Robots}\label{subsec:mujoco}
\begin{figure}[t]
    \centering
    \includegraphics[width=\linewidth]{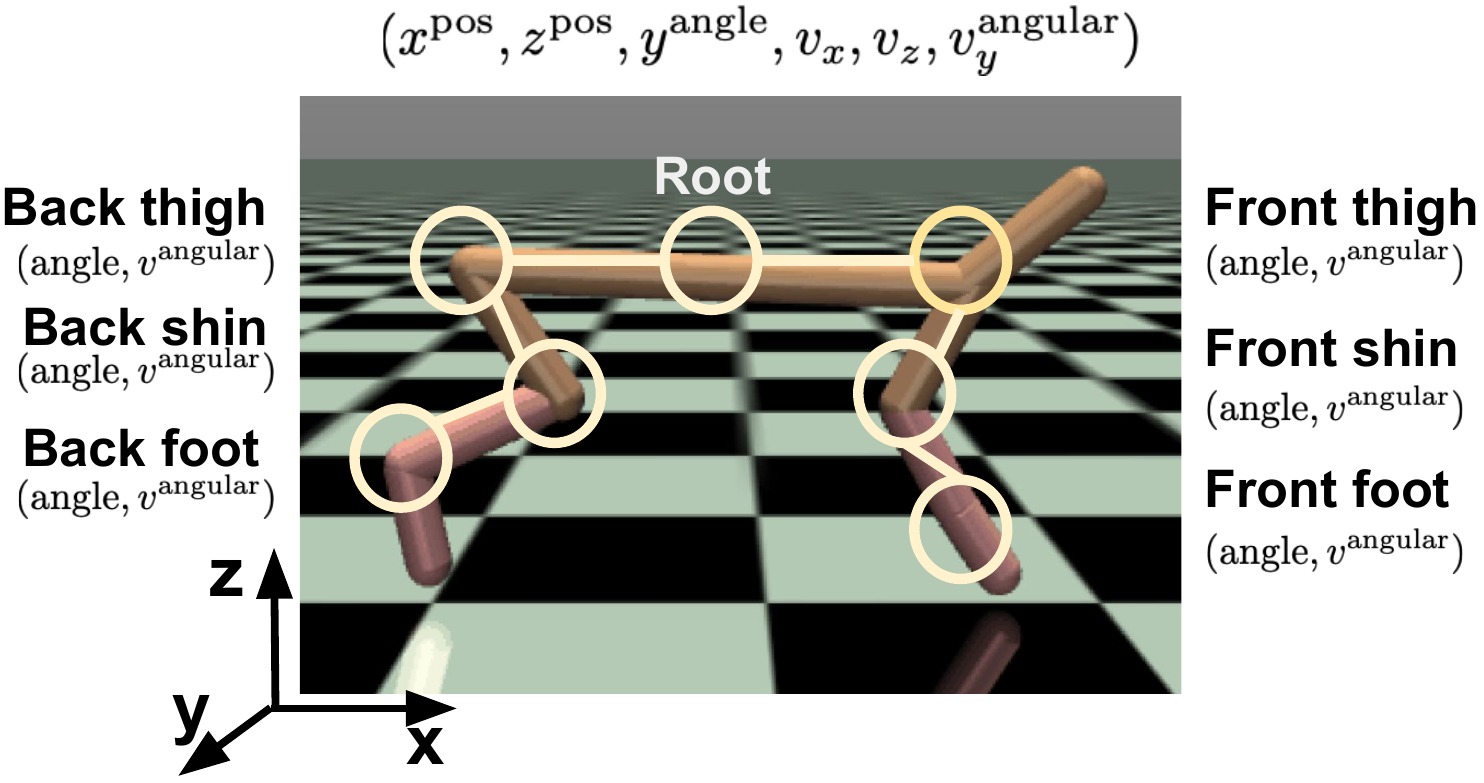}
    \caption{Multiagent MuJoCo Cheetah}
    \label{fig:mujoco}
\end{figure}

A robot capable of performing continuous control can be defined via a kinematics tree~\cite{spong2006robot, wang2018nervenet} of nested bodies, specifying the mechanical structure and physical properties of the robot. Within such a kinematics tree, there are three major types of entities: \texttt{Bodies, Joints and Actuators}. The \texttt{Bodies} represent the rigid bodies of the robot, e.g., foot. Physical properties such as the relative position and geometry are specified for each body. The \texttt{Joints} are moveable components which connect parent and child bodies, and creates motion degrees of freedom between them. Typical joints include ball joints (3 rotational degree of freedom), hinge joints (1 rotational degree of freedom) and slide joints (1 translational degree of freedom). For example, in a cheetah robot (Figure~\ref{fig:mujoco}), the torso (root \texttt{body}) can perform translational and rotational movements through slide and hinge \texttt{joints}. The back foot (\texttt{body}) is attached to its parent back shin (\texttt{body}) through a hinge \texttt{joint}. Attached to the joints are \texttt{Actuators} which are the components that actuate the joints. Many real life robots consist of multiple connected links,  
we now introduce how to model the robot as a multiagent system. 


\section{Multiagent Continuous Control}\label{sub:maloco}

\subsection{Problem Formulation}\label{subsec:robot_multiagent}

\noindent \subsubsection{Robot Joints as a Multiagent System} Figure~\ref{fig:mujoco} shows an example cheetah robot built using MuJoCo~\cite{brockman2016openai} (more details in Section \ref{exp:mujoco}).
We model the robot as a multiagent system of $N$ agents, where each agent represents: 1. a rigid body component (e.g., the back foot) of the robot, 2. the joint which attaches the body to its parent (e.g., the ankle joint which attaches the back foot to the back shin), and 3. any actuators attached to the joint. In this way, the agent receives input of the local sensor observations to its joint (e.g., positions, angles, translational and angular velocities, external forces and torques), and outputs a control action for the actuator. 

\noindent \subsubsection{Multi-agent based Robot Locomotion.} Typically in the robotic control problems, the actuators have a continuous action space which are normalised to $[-1,1]$ for the convenience of learning. In this work, we focus on the robot locomotion task, which is a representative continuous robot control problem in this family. 
The objectives of robot locomotion tasks are to train the robot to learn control policies in order to transport from place to place through walking, hopping, swimming, etc. Here we model the locomotion tasks as a fully cooperative Multiagent decision making problem, described by a Markov game~\cite{littman1994markov} denoted by a tuple $\langle N, \states, \actions, \dynamics, \reward, \gamma \rangle$, where $N$ is the set of all agents $N=\{1,\ldots, n\}$, and $\states$ are global states of the environment. Each agent $i$ has a continuous action space $\actions_i$ and the combined action space of the robot is $\actions = \actions_1 \times \ldots \times \actions_n$. At each step $t$, each agent $i$ makes a local observation $s^i_t = o^i(s_t)$, and chooses an action simultaneously according to its own policy $a^i_t\sim\pi^i(s^i_t)$. Then, a combined action composed of the actions of all agents $\mathbf{a}_t = (a^1_t, \ldots, a^n_t)$ is performed on the environment (the combined policy of all agents is denoted as $\pi = (\pi^1,\ldots, \pi^n)$). Upon taking the action, all agents receive a shared reward $r_t = \reward(s_t, \mathbf{a}_t)$ which evaluates how well the robot performed (e.g., the distance travelled along a specified direction, the control energy consumption, the stability of the robot, etc), and the environment transitions to the next state according to the transition dynamics $\dynamics:\states\times\actions\rightarrow\states$. At each step, the agents aim to maximise the cumulative future discounted return $G_t = \sum_k{\gamma^k}r_{t+k+1}$ where $\gamma\in [0, 1]$ is a discount factor.

\subsection{Multiagent PPO with Shared Advantage}\label{sub:mappo}

Having formulated the multiagent decision problem, we next introduce an algorithm for optimising the agents' policies.
To optimise the continuous control policies, we extend the standard PPO algorithm with actor-critic framework (c.f. Section~\ref{subsec:PPO}) to cooperative multiagent PPO. In particular, we adopt the centralised training, decentralised execution (CTDE) paradigm~\cite{foerster2018counterfactual}, c.f. Section~\ref{subsec:CTDE}. As the agents are fully cooperative and the whole robot receives a global reward signal, we let the agents share a centralised critic which estimates the value function of the global state $V_\vparam: \states\rightarrow \mathbb{R}$. The centralised critic is parametrised by $\vparam$, and in our case, $\vparam$ is a neural network. The critic is only used during training as a guide for optimising the actors and will be dismissed in the execution phase---during the execution only the actors are used to choose the actions. The actors are decentralised, i.e., for the $i-$th agent, its actor $\pi_{\aparam^i}: \states^i \rightarrow \actions^i$ parametrised by neural network $\aparam^i$, chooses its own action given the local observations. Most locomotion tasks have continuous action space by nature; hence we model each agent's policy outputs by a Gaussian distribution. At each step, action is sampled from the Gaussian distribution. During training, each actor learns to update the mean of the Gaussian distribution, and the standard deviation can be either static or learned. 

We optimise the centralised critic and decentralised actors following the PPO algorithm: 
At each epoch $k$, the agents collect a set of trajectories $\data_k$ by running their current policy $\pi_k = (\pi^1(\aparam^1_k), \ldots \pi^n(\aparam^n_k))$ in the environment.
Then, the critic which estimates the (state) value function $V(s)$ is updated by regression. The loss is defined as the mean squared error between the predicted value of the state $s_t$ the empirical return $\hat{G}_t$ from $t$ onwards:
\begin{equation*}
	\vparam \leftarrow \arg\min_\vparam \frac{1}{|\data_k|T}\sum_{\tau\in{\data_k}}\sum_{t=0}^T\bigl(V_\vparam(s_t) - \hat{G}_t\bigr)^2.
\end{equation*}
The actor of each agent is updated by optimising the PPO clipped surrogate objective: 
\begin{align*}
\aparam^i_{k+1} = \arg\max_{\aparam^i} \frac{1}{|\data_k|T} \sum_{\tau\in\data_k}\sum_{t=0}^T \min\Big(\frac{\pi_{\aparam^i}(a^i_t|s^i_t)}{\pi_{\aparam^i_k}(a^i_t|s^i_t)} A(s_t, \mathbf{a_t}),\\ \textnormal{clip}\bigl(\frac{\pi_{\aparam^i}(a^i_t|s^i_t)}{\pi_{\aparam^i_k}(a^i_t|s^i_t)}, 1-\epsilon, 1+\epsilon\bigr)A(s_t, \mathbf{a_t}) \Big),
\end{align*}
\noindent where $A(s_t, \mathbf{a}_t)$ is the shared advantage function computed using the critic through generalised advantage estimation (c.f. Sec.~\ref{subsec:PPO}). Intuitively, if the advantage $A(s, a)$ is positive, the actor parameters update in the direction where this action becomes more likely. PPO avoids parameter updates that change the policy too much by clipping the ratio between old and new policies~\cite{schulman2017proximal}.

\begin{figure*}[t]
    \centering
    \includegraphics[width=\linewidth]{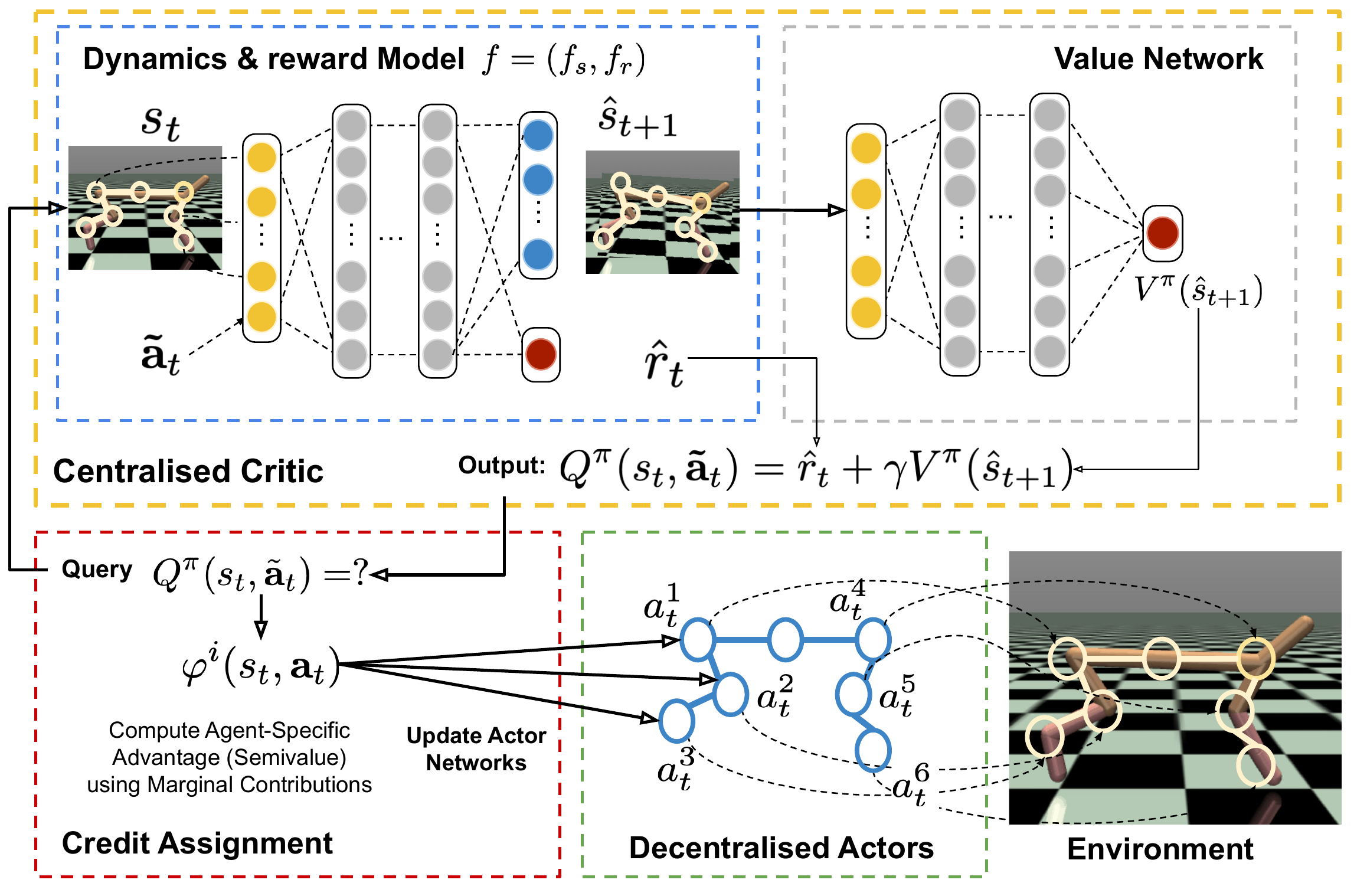}
    \caption{Model-based Multiagent Credit Assignment. Our framework consists of three modules: (yellow) the centralised critic module which consists of a dynamics/reward model $f=(f_s, f_r)$ and a centralised state-value function $V^\pi(s)$; (red) credit assignment module which queries the critic for coalition values and computes the counterfactual solution concepts assigned to each agent; (green) actors module with $N$ agents, which are updated using the credits from the credit assignment module.}
    \label{fig:mb}
\end{figure*}

\subsection{Multiagent PPO with Game-theoretic Agent-specific Advantage}\label{sub:gt}

The above Multiagent PPO with shared advantage function often yields decent performance for cooperative multiagent tasks. However, the shared advantage function $A(s_t, \mathbf{a}_t)$ only evaluates the quality of the combined action, which fails to assess each agent's individual contribution. As we will discuss in Sec.~\ref{exp:central_vs_multi}, failing to address per-agent contribution can result in low sample efficiency oftentimes, for example, an under-actuated agent can equally share and update its policy using the global advantage. To deal with this issues, we leverage the fair credit assignment methods from cooperative game theory, and present a generic game theoretic framework for agent-specific advantage computation in Multiagent PPO. 

\subsubsection{The Characteristic Function}\label{subsec:mc}
To compute the value of an agent assigned by the game-theoretic solution concepts, we first need to define the characteristic function. At each timestep $t$, given the environment state $s_t$ and agents' joint action $\mathbf{a}_t = (a^1_t,\ldots, a^n_t)$, let the value of a coalition $\coalition$ be defined as:
\begin{align}
    \cv^\coalition( s_t, \mathbf{a}_t) &= Q^\pi(s_t, \Tilde{a}^1_t, \ldots, \Tilde{a}^n_t), \textnormal{ where}\label{eq:coalition}\\
    \Tilde{a}^i_t &= 
        \begin{cases}
            a^i_t \quad \quad \quad \textnormal{if } i\in\coalition\\
            a_\textnormal{default} \quad \textnormal{if } i\in N\setminus\coalition,
        \end{cases}\nonumber
\end{align}
\noindent where $\Tilde{a}^i_t$ denotes the action of agent $i$ is replaced by a default one if $i$ is outside the coalition, a widely adopted practice~\cite{li2021shapley,wolpert2002optimal}. To follow the game theoretic conventions where empty coalitions have zero value, we can normalise the characteristic value function by subtracting a baseline value $Q^\pi(s_t, \mathbf{a}_\textnormal{default})$. However, the marginal contributions will stay invariant so we will use the above definition for simplicity. Given this characteristic function, agent $i$'s marginal contributions towards coalition $\coalition$:
\begin{align}
    \mc^i(\coalition, s_t, \mathbf{a}_t) &= \cv^{\coalition \cup \{i\}}(s_t, \mathbf{a}_t) - \cv^{\coalition}(s_t, \mathbf{a}_t),
\end{align}
\noindent represents the difference made by $i$'s chosen action towards coalition $\coalition$, compared with the default action.

\subsubsection{Agent-Specific Advantage}\label{subsec:semi}
We now introduce a generic game-theoretic credit assignment framework for multiagent continuous control using semivalues~\cite{dubey1981value} -- a wide family of solution concepts that encompass many common credit assignment methods such as the Shapley value, Banzhaf value, etc.


\begin{equation*}
	\val^i(\cv) = \sum_{\coalition\subseteq N\setminus\{i\}} \weight{c}{N} \mc^i(\coalition) \textnormal{ where } c=|\coalition|, \sum_{c=0}^{|N|-1}\weight{c}{N}\tbinom{|N|-1}{c} = 1.
\end{equation*}
To better understand them, we can rewrite the semivalues as:
\begin{equation}
	\val^i(\cv) = \sum_{c=0}^{|N|-1} p_c \frac{\sum_{\coalition\subseteq N\setminus\{i\}, |\coalition| =c} \mc^i(\coalition)}{\tbinom{|N|-1}{c}}, \textnormal{where} \sum_{c=0}^{|N|-1}p_c = 1.
	\label{eq:semivalue}
\end{equation}

Note that the denominator $\tbinom{|N|-1}{c}$ is the number of size-$c$ coalitions excluding player $i$. Intuitively, the semivalue is a weighted sum of an agent's average marginal contribution towards different sized coalitions, 
where $p_c$ form a probability distribution. In particular, 
the Shapley value places uniform weight on all coalition sizes: $p_c = \frac{1}{|N|}$;
and the Banzhaf value has a bell-shaped distribution, i.e., $p_c  = \frac{1}{2^{|N|-1}}\tbinom{|N|-1}{c}$. Any other probability distribution $p_c$ also specifies a semivalue.

In our model, \emph{at each timestep, the semivalue $\val^i$ of an agent computes the contributions of the agent's chosen action towards the global future return of the robot, and hence will be used as the agent-specific pseudo advantage for multiagent PPO updates.} 
We call the semivalues ``pseudo'' advantage because they are not defined strictly as $A(s, a) = Q(s, a) - V(s)$. Nevertheless, the semivalues provide an effective proxy of advantage as they can evaluate the individual contribution of an agent's action towards the global value. 

Since RL training procedures typically require millions of timesteps, enumerating all possible coalitions and computing their characteristic value at each time step is inefficient. 
A simple procedure for estimating the semivalue is through Monte-Carlo sampling and output the average over the agent's marginal contribution towards all sampled coalitions. For each sample drawn, we can first sample a coalition size $c_m$ according to the semivalue distribution $p_c$, then uniformly sample a coalition $\coalition_m$ of the size $c_m$. 

\subsection{Model-based Advantage Estimation}\label{sub:mb}
The semivalues (e.g., the Shapley and Banzhaf values) as agent-specific pseudo advantages, allow us to evaluate the contribution of each agent and its chosen action. 
\emph{However, how can we obtain the coalitions' values, when only the value of the grand coalition is available (i.e., the global reward of the whole system)?} 

An obvious approach is to perform extra simulations, where for each simulation the grand coalition is replaced by a different coalitions of agents. This, however, would require exponentially more simulations which is expensive for mutliagent RL domains in general. Another approach which does not require extra simulations is to infer the value of coalitions from the present simulations~\cite{li2021shapley}. However, this approach suffers from inaccurate estimations in the multiagent robotics control domain, due to the high dimensional continuous state and action spaces.

To accurately estimate the value of different coalitions, we draw inspiration from model-based RL and propose incorporating an additional model of the environment dynamics and rewards. In this way, we obtain a better coalition value estimation through model-based simulations of different coalitions. Meanwhile, optimisation of the model is straightforward (via supervised learning) and does not require extra environment interactions.





\begin{algorithm}[t]
	\caption{Model-based Multiagent Credit Assignment}
	\label{algo:mb}
	\begin{algorithmic}[1]
		\State{\textbf{Input:}  $N=\{1,\ldots,n\}$: the set of all agents.}
		\State{\textbf{Initialise: } model $f = (f_s, f_r)$, centralised critic $V_{\vparam}$, decentralised actors $\pi^i_{\aparam}$ for each agent $i\in N$}
		\For{iterations $k=0,1,2,\ldots $} 
		    \State{Collect sets of trajectories $\data_k=\{\tau\}$ by running policy $\pi_k = (\pi^1_{\aparam^1_k}, \ldots \pi^n_{\aparam^n_k})$ in the environment. Compute returns $\hat{G}_t$}
		    \State{$\#$ Fit the dynamics/reward model (in minibatches):}
            \State{\phantom{$\#$} \resizebox{.9\hsize}{!}{$f_s \leftarrow  \arg\min_{f_s} \frac{1}{|\data_k| T}\sum_{\tau \in \data_k}\sum_{t=0}^{T}\|s_{t+1} - (s_{t} + f_s(s_t, \mathbf{a}_t))\|^2$}}
            \State{\phantom{$\#$} \resizebox{.8\hsize}{!}{$f_r \leftarrow \arg\min_{f_r} \frac{1}{|\data_k|T}\sum_{\tau \in \data_k}\sum_{t=0}^{T}\|(r_t - f_r(s_t, \mathbf{a}_t)\|^2$}}
            \State{$\#$ Compute per-agent advantage $\val_i$ for all timesteps $t$:}
            \State{\phantom{$\#$} Sample coalitions $\coalition_m$}
            \State{\phantom{$\#$} Compute coalition values by model $f$ and critic $V_{\vparam_k}$: }
            \State{\phantom{$\#$}\phantom{----}$v(C_m, s_t, \mathbf{{a}}_t)) = f_r(s_t, \mathbf{\Tilde{a}}_t) + \gamma V^\pi_{\vparam}(f_s(s_t, \mathbf{\Tilde{a}}_t) + s_t)$}
            
            \State{\phantom{$\#$} Compute marginal contribution $\mc^i(\coalition_m, s_t, \mathbf{a}_t)$} 
            \State{\phantom{$\#$} Compute semivalues $\val^i$ using the marginal contributions. }
            \State{$\#$ For each agent $i$, update the policy by maximising the PPO-Clip objective:}
            \State{\phantom{$\#$} $\aparam^i_{k+1} = \arg\max_{\aparam^i} \frac{1}{|\data_k|T} \sum_{\tau\in\data_k}\sum_{t=0}^T \min\Bigl(\frac{\pi_{\aparam^i}(a^i_t|s^i_t)}{\pi_{\aparam^i_k}(a^i_t|s^i_t)}\val^i(s_t, \mathbf{a_t}),$} \State{\phantom{ space}$\textnormal{clip}\bigl(\frac{\pi_{\aparam^i}(a^i_t|s^i_t)}{\pi_{\aparam^i_k}(a^i_t|s^i_t)}, 1-\epsilon, 1+\epsilon\bigr)\val^i(s_t, \mathbf{a_t}) \Bigr)$,}
            \State{$\#$ Fit the centralised critic by regression:}
            \State{\phantom{$\#$} $\vparam_{k+1} = \arg\min_\vparam \frac{1}{|\data_k|T}\sum_{\tau\in{\data_k}}\sum_{t=0}^T\bigl(V_\vparam(s_t) - \hat{G}_t\bigr)^2$}
		\EndFor
	\end{algorithmic}
\end{algorithm}
\subsubsection{Dynamics and Reward Model}\label{subsec:model}
We use a \emph{model} parametrised by $\mparam = (\mparam_s, \mparam_r)$, which consists of a transition model $\tmodel$ which maps the state-action tuple to the the difference between the next and current state (alternatively, $\tmodel$ can map the state-action tuple to the next state directly), and a reward model $\rmodel$ which maps the state-action tuple to the reward~\cite{nagabandi2018neural}:
\begin{align}
    \forall s_t\in \states, a_t\in \actions, \quad
    \hat{s}_{t+1} &= \tmodel(s_t, a_t) + s_{t},\\
    \hat{r}_{t} &= \rmodel(s_t, a_t). 
\end{align}
The model can be optimised through supervised learning, by regression using mean-squared error between the predicted next states/rewards and actual ones.
\begin{align}
    &\min_{\mparam_s} \frac{1}{|\data|}\sum_{\langle s_t, s_{t+1}, \mathbf{a}_t, r_t\rangle\in \data}\|s_{t+1} - (s_{t} + f_s(s_t, \mathbf{a}_t))\|^2\\
    &\min_{\mparam_r} \frac{1}{|\data|}\sum_{\langle s_t, s_{t+1}, \mathbf{a}_t, r_t\rangle\in \data}\|(r_t - f_r(s_t, \mathbf{a}_t)\|^2.
\end{align}

The model is only used in (centralised) training, hence only one centralised model is needed and can compensate for the state of the whole system. During each iteration of training, we alternate between model fitting and agent updates using the data collected from environment interactions. Moreover, we only use the model for generating imaginary simulations of short-horizons, which mitigates the error-prone model predictions drifted in long horizons.

\subsubsection{Estimating the coalition values using the model}

Inspired by model-based value expansion~\cite{feinberg2018model}, we use the model to perform estimation of the one-step transition dynamics and rewards. In this way, we obtain a better value estimation through imaginary simulations of different coalitions using the dynamics/reward model. Through the Bellman equation, we can estimate the action-value function through a state-value function $V^\policy$ in the following:
\begin{align}
    Q^\policy(s_t, \Tilde{\mathbf{a}}_t) 
    &= \mathbbm{E} [r_t + \gamma V^\policy(s_{t+1}) |\phantom{s} s_t, a_t ] \nonumber\\
    &= \mathbbm{E} [\hat{r}_t + \gamma V^\policy(\hat{s}_{t+1}) |\phantom{s} s_t, a_t ]\nonumber\\
    &\approx f_r(s_t, \Tilde{\mathbf{a}}_t) + \gamma V^\policy(s_t + f_s(s_t, \Tilde{\mathbf{a}}_t)).
\end{align}

A schematic of the framework is shown in Figure~\ref{fig:mb}. At each training step, to produce the credit value for each agent $i$, the credit assignment module first samples coalitions $\coalition_m\subseteq N\setminus\{i\}$ according to the semivalue, and computes the average over counterfactual value of $i$ towards the sampled coalitions $\mc_i(\coalition_m) = \cv(\coalition_m\cup\{i\}) - \cv(\coalition_m)$. To obtain the value of the coalitions, say $\cv^{\coalition_m}(s_t, \mathbf{\Tilde{a}}_t)=Q^\pi(s_t, \Tilde{\mathbf{a}}_t)$, the credit assignment module queries the critic module (yellow in Figure~\ref{fig:mb}): the model (blue) first produces the next state $\hat{s}_{t+1} = f_s(s_t, \Tilde{\mathbf{a}}_t)$ and reward $\hat{r}_t = f_r(s_t, \Tilde{\textbf{a}}_t)$, then the centralised state-value critic (grey) produces $V^\pi(\hat{s}_{t+1})$. Together with the reward $r_t$ produced by the model, the critic module outputs the value of the coalition $\coalition_m$ as $Q^\pi(s_t, \Tilde{\textbf{a}}_t) = \hat{r}_t + \gamma V^\pi(\hat{s}_{t+1})$. Finally, having obtained the credit value assigned to each agent, the actors module (green) can update the decentralised actors through multiagent PPO.
The full algorithm is shown in Algorithm~\ref{algo:mb}.

\section{Experiments}

\begin{figure*}[t]
	\centering
	\begin{minipage}{1\linewidth}\centering
		\includegraphics[height=0.04\linewidth]{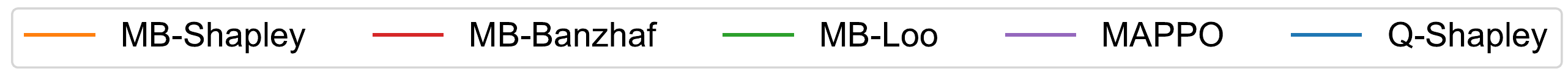}
	\end{minipage}
	\vfill
	\vspace{-3mm}
	\subfloat[Cheetah (Rewards)]{\includegraphics[width=0.47\linewidth]{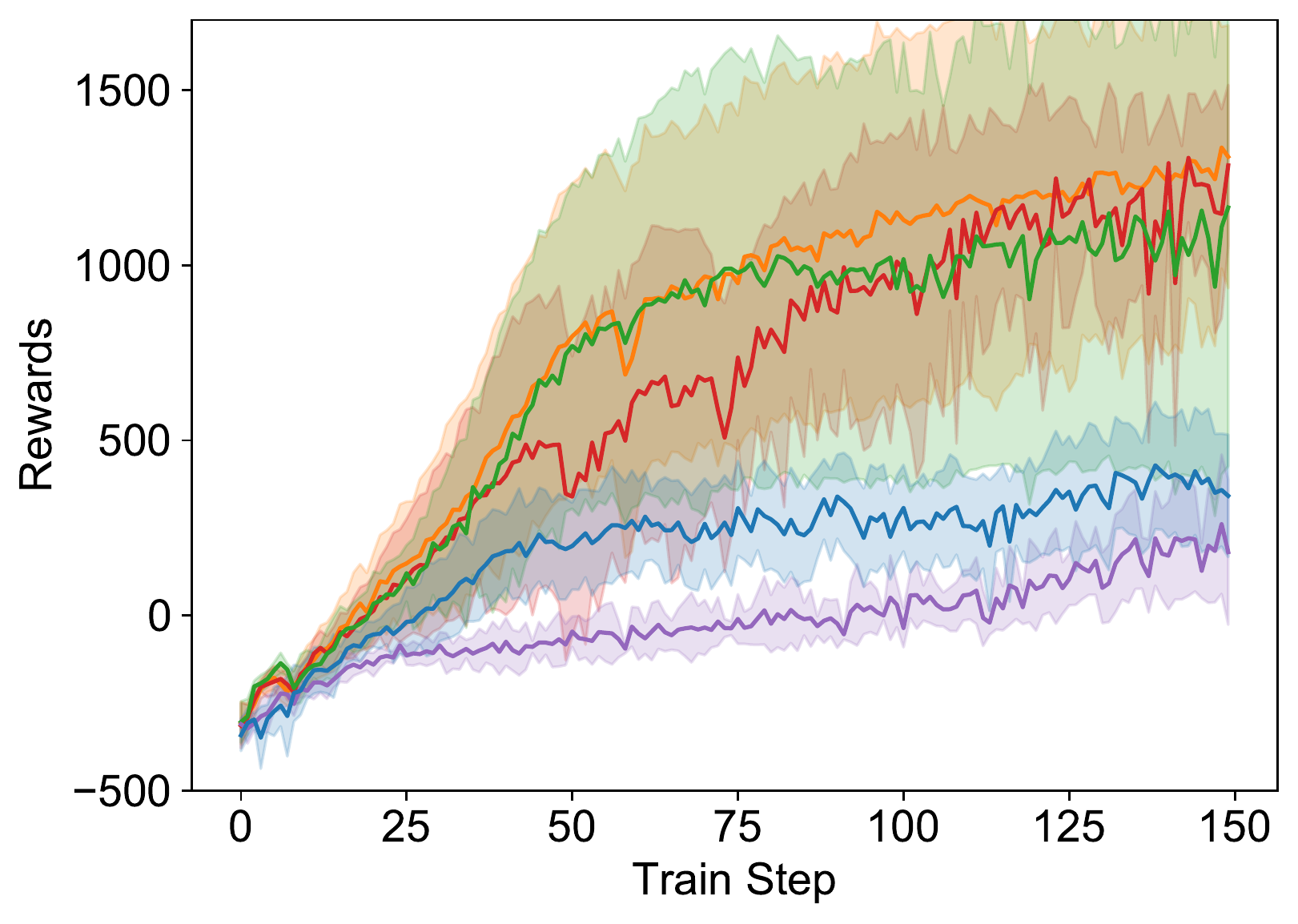}\label{fig:cheetah}}%
	\hfill
	\subfloat[Ant (Rewards)]{\includegraphics[width=0.47\linewidth]{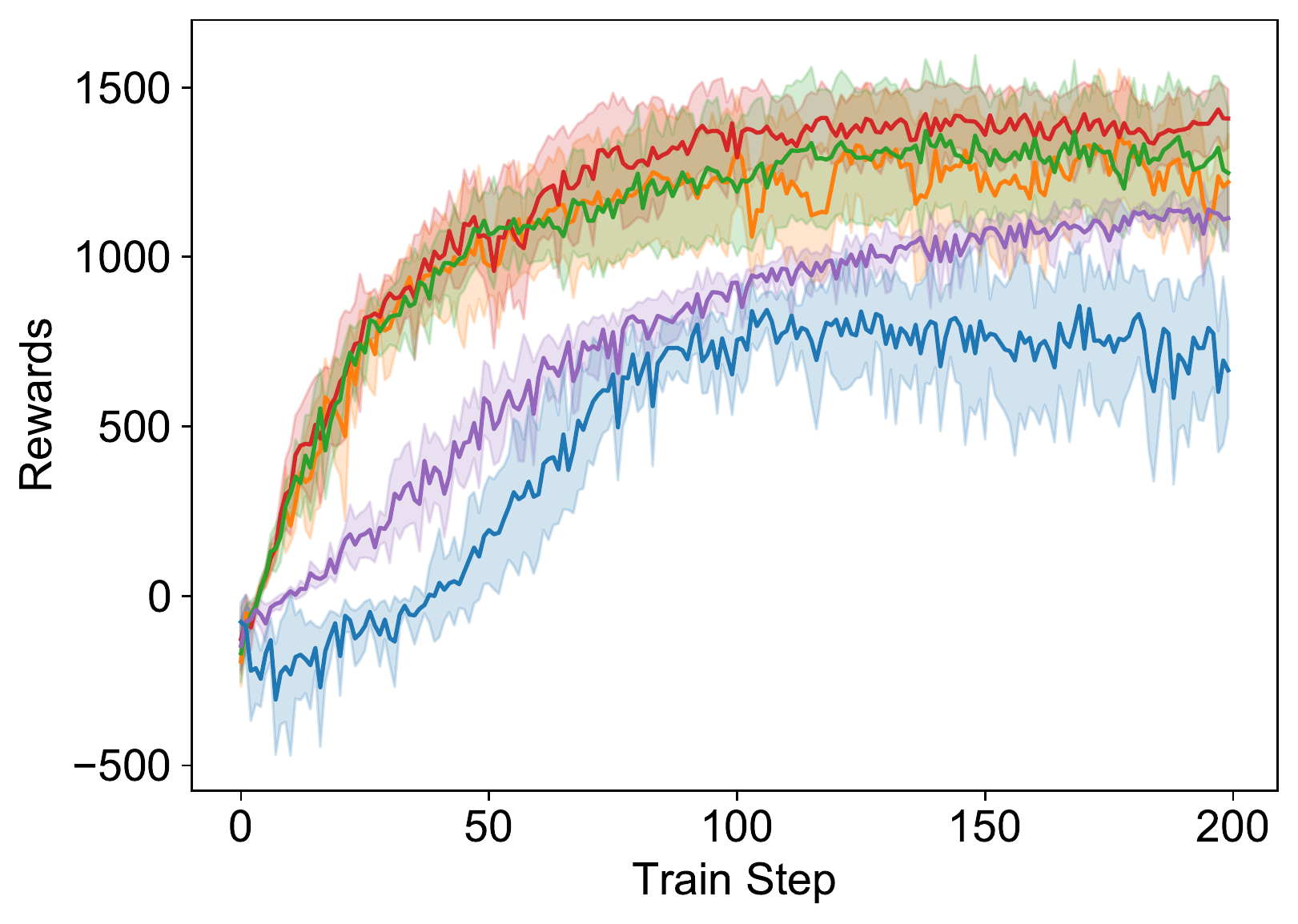}\label{fig:ant}}%
	\vfill
	\vspace{-3mm}
	\caption{Average Rewards Multiagent Cheetah and Ant}
	\label{fig:all_results}
\end{figure*}

\begin{figure}
\centering
	\begin{minipage}{1\linewidth}\centering
		\includegraphics[height=0.05\linewidth]{img/legend.pdf}
		\vspace{-1cm}
	\end{minipage}
	\vfill
	\subfloat[Cheetah (Average Actions)]{\includegraphics[width=0.5\linewidth]{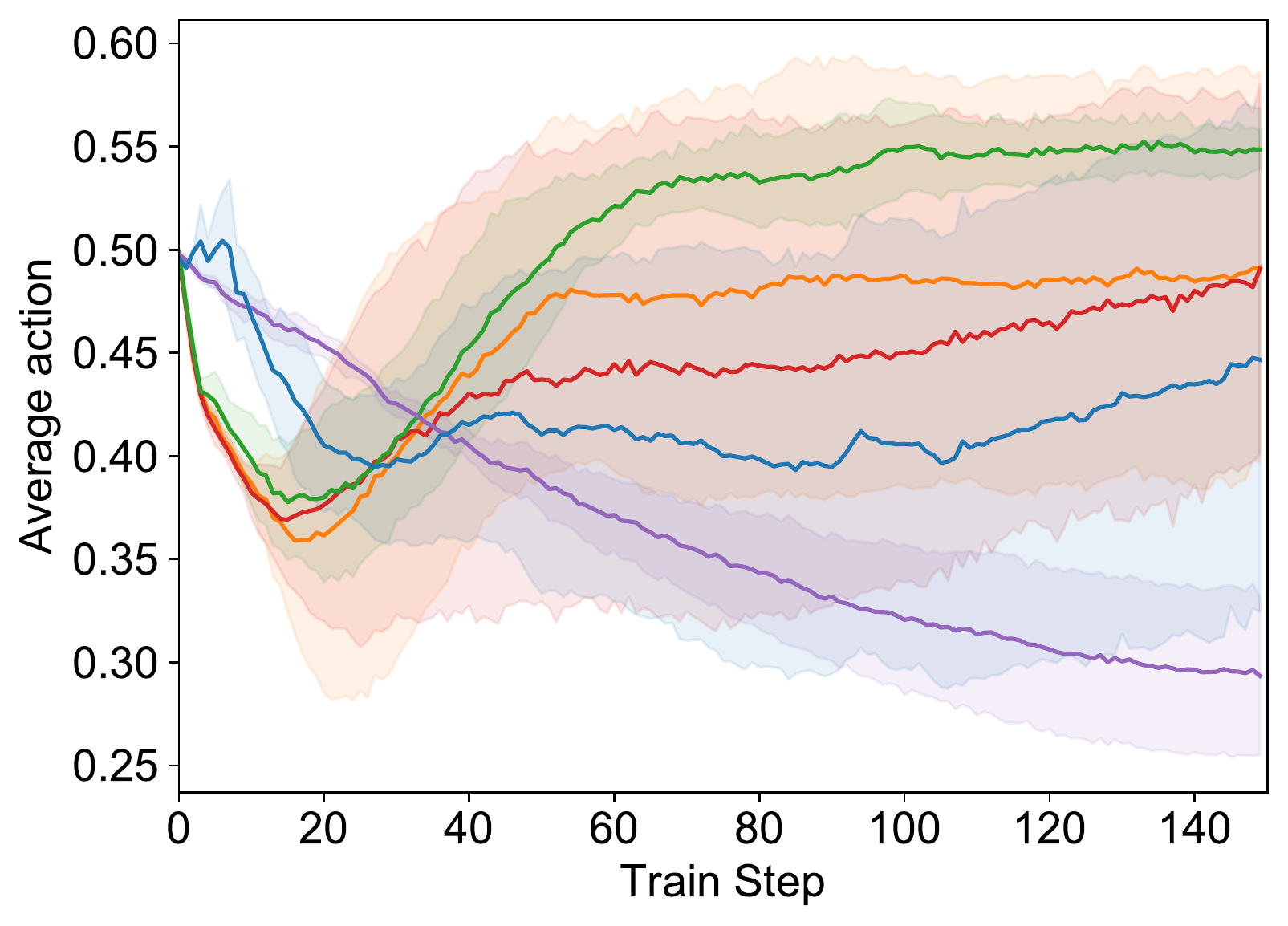}\label{fig:cheetah_actions}}%
	\hfill
	\subfloat[Ant (Average Actions)]{\includegraphics[width=0.5\linewidth]{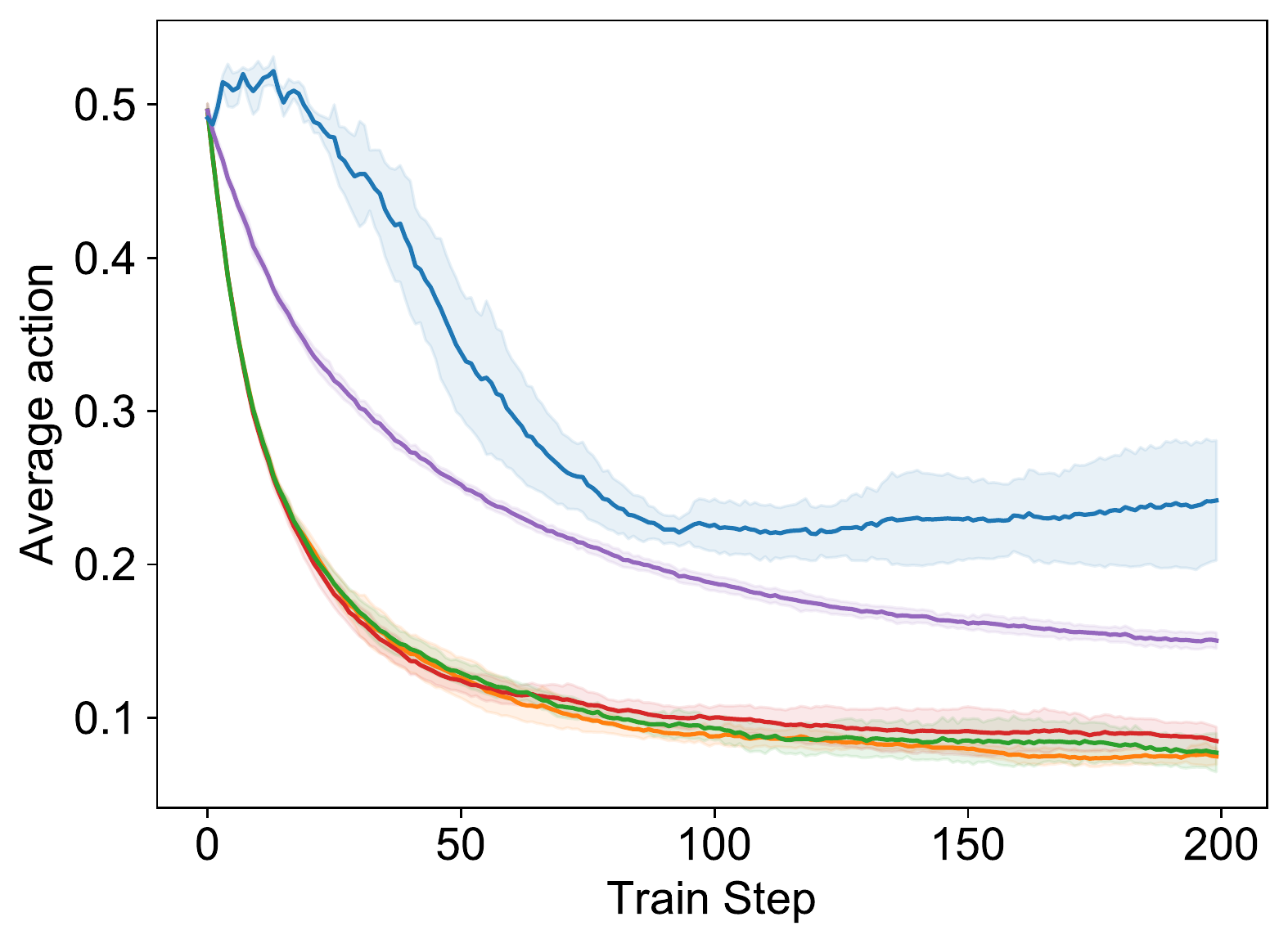}\label{fig:ant_actions}}%
	\caption{Average Actions Multiagent Cheetah and Ant}
	\vspace{-5mm}
\end{figure}

In this section, we present empirical results of our model-based multiagent RL framework on continuous tasks.

\subsection{Experiment Setups}
\subsubsection{Experiment Settings}\label{exp:mujoco}
For the experiments we use OpenAI Gym MuJoCo~\cite{brockman2016openai, todorov2012mujoco} locomotion tasks, which are the standard benchmarks for continuous control with RL~\cite{wang2018nervenet, nagabandi2018neural}. 
Two exemplar robot locomotion tasks are selected as our case studies: (1) the Cheetah movement in 2D space and (2) the Ant movement in 3D space. The cheetah model in MuJoCo has 7 joints, including one root joint and 6 joints each paired with one actuator.  Following the definition in Section~\ref{subsec:robot_multiagent}, we consider each joint as an agent with an actuator that applies a joint torque. For all agents, the action space is normalised to $[-1, 1]$. The root agent observes the position, angle, velocity and angular velocity and for other agents, their observation includes the relative angle with respect to the body that they are attached to, and the angular velocity. The global observation utilised by the critic and the dynamics model (c.f. Figure~\ref{fig:mb}) is given by a concatenation of local observations. The objective of the agent is to move forward in the $x$-direction. On the other hand, the Ant model can move in a 3D space and the agents are similarly defined as in the Cheetah. Notably, the agents of an Ant model in MuJoCo can additionally observe external forces such as friction. The objective of the Ant locomotion task is also to move forward in the $x$-direction.

\subsubsection{Models and Training Details}
Our multiagent system consists of a model $f = (f_s, f_r)$ of the environment, the centralised critic, and the decentralised actors.
We use PyTorch~\cite{paszke2019pytorch} for implementing and optimising the neural network models. For the dynamics model $f_s$, we use 4 fully-connected layers with ReLU activation function~\cite{nwankpa2018activation} and a linear output layer with the dimension of $128$. For the reward model $f_r$, we use 3 fully-connected layers with ReLU activation function and a linear output layer with the dimension of $128$. 
The model is optimised in minibatches of size $64$ using Adam optimiser~\cite{kingma2014adam} with the learning rate of $10^{-3}$. For the centralised critic, we use $3$ fully-connected layers of the dimension of $32$ with Tanh~\cite{nwankpa2018activation} activation and a linear output layer which produces the value of the input state. For the decentralised actors, each actor uses 3 fully-connected layers of the dimension of $32$ with Tanh activation, and an output layer with Tanh activation. The default actions $a_\textnormal{default}$ are given by zero vectors. Both the critic and the actor are optimised using the Adam optimiser. The learning rate for the critic is $10^{-3}$ and is $3\times 10^{-4}$ for the actor. All graphs plot mean and standard deviation across 5 seeds. 

\subsection{Overall Results}
Figure~\ref{fig:all_results} shows the overall performance of our algorithms and baseline algorithms on the MuJoCo locomotion tasks with cheetah and ant robots. The $y$-axis shows the episode rewards and $x$-axis shows the train steps. 
In particular, we follow Eq.~\ref{eq:semivalue} to implement the following variants of our semivalue based credit assignment algorithms. \texttt{MB} is for Model-based coalition value estimation:
\begin{itemize}
	\item{\mbshapley: the Shapley value ($p_c=\frac{1}{|N|}$);} 
	\item{\mbbanzhaf: the Banzhaf value  ($p_c=\frac{1}{2^{|N|-1}}\tbinom{|N|-1}{c}$);}
	\item{\mbloo: the Leave-one-out value ($p_c=\mathbbm{1}_{c=|N|-1}$), where we compute the marginal contribution of an agent towards the grand coalition (made of all other agents);}
\end{itemize}
For \emph{baselines}, we compare with the following state-of-the-art algorithms in multiagent RL that use the Shapley value-based credit assignment and the global advantage respectively:
\begin{itemize}
	\item{\qshapley~\cite{li2021shapley}: we adapt the discrete domain algorithm~\cite{li2021shapley} which used Shapley value for credit assignment, to our multiagent PPO which can be used for continuous control. The coalition values are computed using a centralised critic of Q-values in a model-free manner.}
	\item{\mappo (Multiagent PPO)~\cite{yu2021surprising}: In \mappo, the centralised critic computes a \emph{shared advantage function (GAE)} for all agents based on the global reward signal. And the agents are trained using centralised training, decentralised execution.}
\end{itemize}

\subsubsection{Shared advantage vs. agent-specific advantage. }\label{exp:central_vs_multi}
We are now in a position to present the results and findings. 
In both the cheetah (Figure~\ref{fig:cheetah}) and ant (Figure~\ref{fig:ant}) cases, we observe that our credit assignment methods that use model-based estimation (i.e., \mbshapley, \mbbanzhaf, \mbbanzhaf) significantly outperform multiagent PPO (\mappo) in terms of both the \emph{average rewards} and \emph{sample efficiency}. For example, in the Ant case, our model-based algorithm with Banzhaf credit assignment (\mbbanzhaf) quickly reaches the reward of 1000 within 50 train steps, while \mappo needs around 150 train steps to reach the same level. \mbbanzhaf also converges to a higher average reward ($\sim1400$) than \mappo ($\sim1100$). 
In all our results, performing credit assignment and evaluating the per-agent contribution (c.f. Sec.~\ref{sub:gt}) is consistently better than agents using a shared advantage function. A demo video is provided in the supplementary.

\subsubsection{Model-based vs. Model-free credit assignment.}
We next compare our \mbshapley with the baseline method \qshapley~\cite{li2021shapley} in order to understand the role of model-based credit assignment. In our \mbshapley, the coalition values are estimated using the dynamics/reward model and the state-value critic $V^\pi$. While in \qshapley, the coalition values are estimated in a model-free way with action-value critic $Q^\pi$. As shown in 
Figure~\ref{fig:cheetah} and~\ref{fig:ant}, for both locomotion tasks, \mbshapley outperforms \qshapley. 
This shows that when only the reward of the grand coalition (i.e., the whole robot) is provided by the simulations, it is difficult to infer the values $\cv^\coalition$ of different coalitions using a model-free Q-value critic. In contrast, a better estimation of values of different coalitions is obtained with the predictions of our model-based module, which enables more effective agent-specific credit assignment and sample-efficient learning. 

\subsubsection{Robustness of Different Semivalue Variants.}
Our model-based coalition value estimation supports the broad class of semivalue credit assignments. Still, from Figure~\ref{fig:all_results}, we observe that the agents trained using different selected semivalues (\mbshapley, \mbbanzhaf, \mbloo) all deliver decent performance. In the case of cheetah (Figure~\ref{fig:cheetah}), both \mbshapley and \mbbanzhaf yield the highest rewards, and in the case of ant (Figure~\ref{fig:ant}), \mbbanzhaf yields the highest rewards. Overall, \mbbanzhaf has the lowest variance among the three variants.

\subsubsection{Average Actions}
Figures~\ref{fig:cheetah_actions} and~\ref{fig:ant_actions} show the mean absolute value of actions, averaged across the agents (joints actuators). This quantity refers to the controller output (normalised between $[0,1]$), and higher action values mean higher energy consumption in a practical context. In cheetah, \mbloo has the highest average action, with \mbshapley and \mbbanzhaf having medium values.  In the ant case, \mbshapley, \mbbanzhaf, and \mbloo all have low control output. We observe that the ant robot trained using \mbshapley, \mbbanzhaf learn to move forward using a subset of joints. In contrast, other joints are mainly learned for steering, implying that using these two semivalues can also encourage diverse roles of different agents in a locomotion task.

\subsection{Component-wise Analysis}

\begin{figure}[t]
	\centering
	\subfloat[Cheetah]{\includegraphics[width=0.5\linewidth]{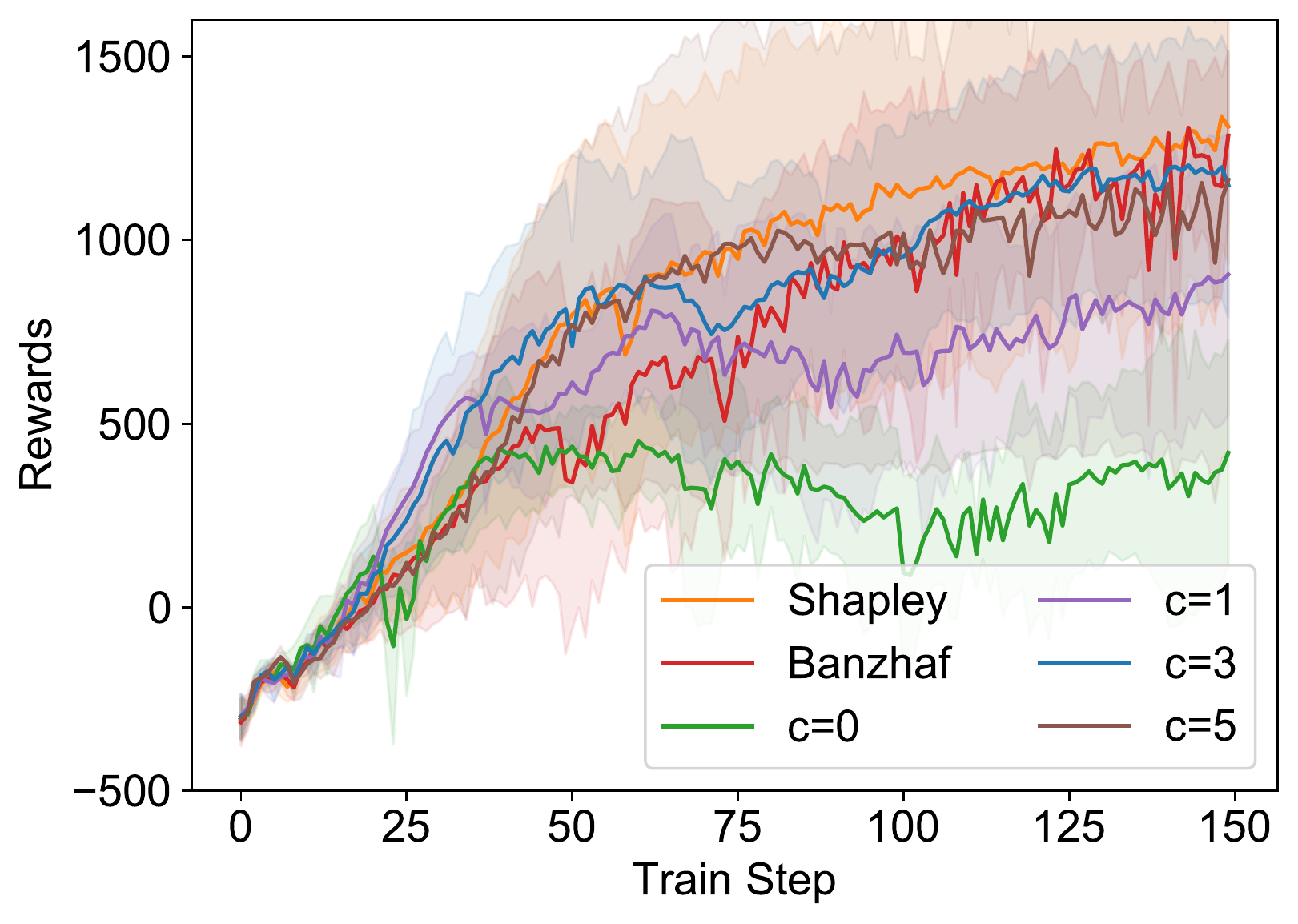}\label{fig:cheetah_size}}%
	\hfill
	\subfloat[Ant]{\includegraphics[width=0.5\linewidth]{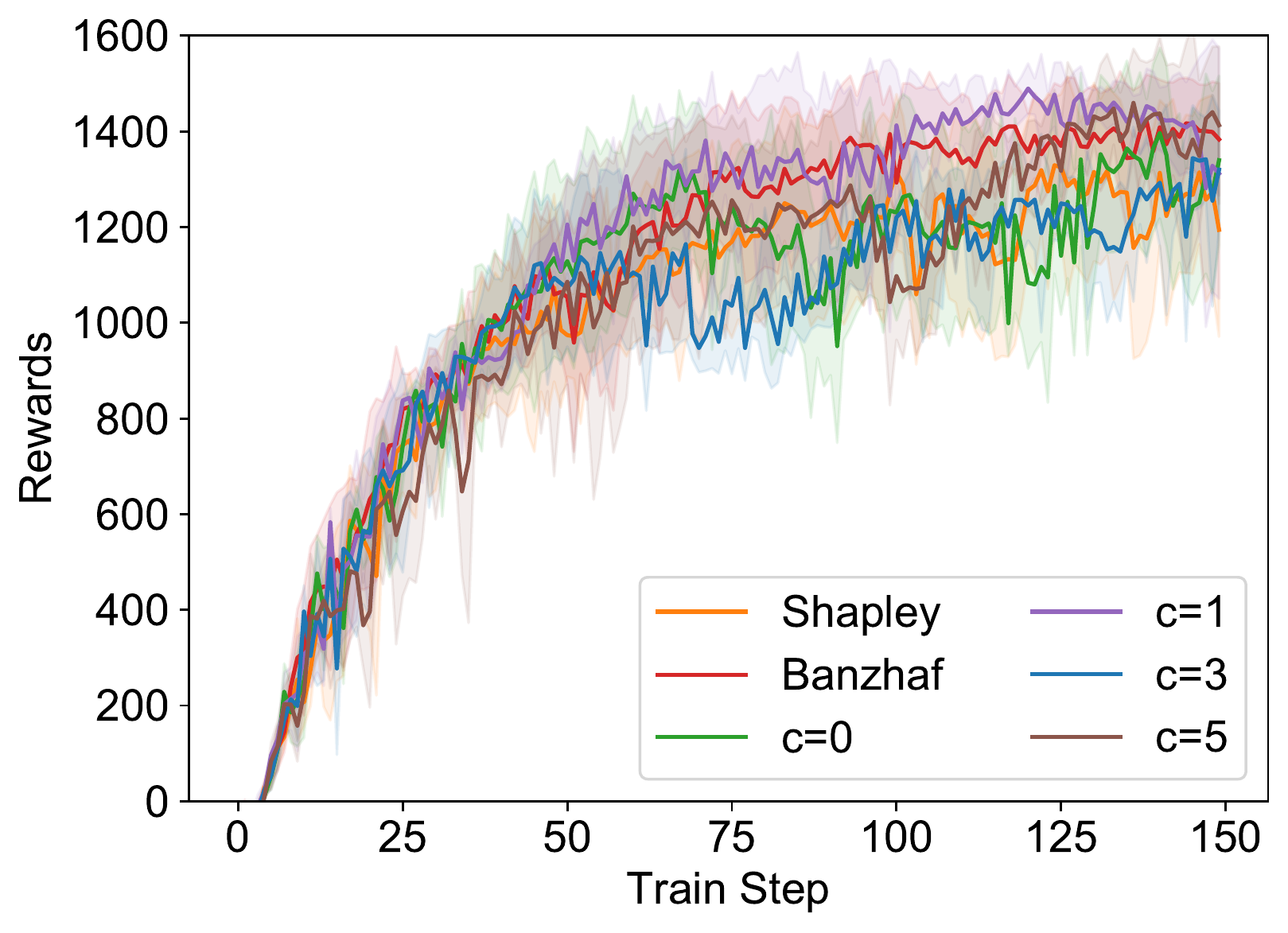}\label{fig:ant_size}}%
	\caption{Results for Varying Coalition Sizes}
\end{figure}
\subsubsection{Variation in Coalition Sizes}\label{subseq:coalition} 
In this section, we discuss the model-based semivalues from the perspective of coalition sizes. Figures~\ref{fig:cheetah_size} and ~\ref{fig:ant_size} show the performance of agents learned using two semivalues \mbshapley ($p_c =\frac{1}{N}$), \mbbanzhaf ($p_c =\frac{1}{2^N}\tbinom{N-1}{c}$), and some semivalues defined by $p_c =\mathbbm{1}_{c=x}$, e.g., a semivalue with $p_c =\mathbbm{1}_{c=3}$ is an agent's average marginal contributions towards size-$3$ coalitions. In the case of cheetah (Figure~\ref{fig:cheetah_size}), 
the semivalues that use small coalition sizes (e.g., $\mathbbm{1}_{c=0}$) yield unsatisfactory performance, while those using mid-sized coalitions (e.g., $\mathbbm{1}_{c=0}$) yield a high reward. In the case of the ant (Figure~\ref{fig:ant_size}), the semivalues focused on
coalitions of heterogeneous sizes all yield high rewards. This suggests that the agents' actions in the cheetah case are more dependent; hence, they require agents to closely coordinate their actions. While for the ant, the agents have more distinct roles and their actions are more independent. The advantage of Shapley value and Banzhaf value is that they take into account all coalitions sizes and therefore are robust across different robots. Shapley value uniformly weighs all coalition sizes while Banzhaf value puts higher weights on mid-sized coalitions.

\begin{figure}[t]
	\centering
	\subfloat[Cheetah]{\includegraphics[width=0.5\linewidth]{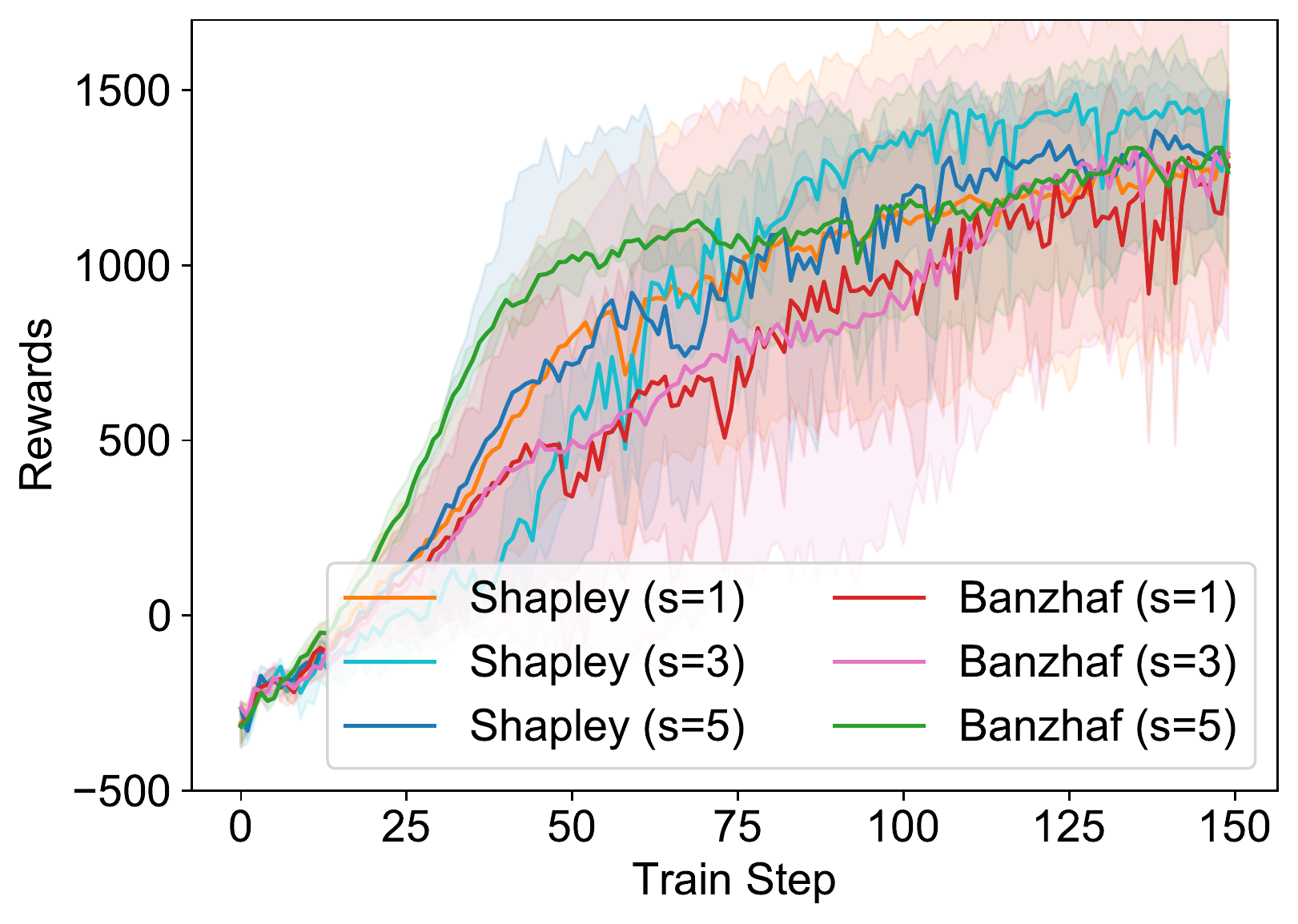}\label{fig:cheetah_sample_size}}%
	\hfill
	\subfloat[Ant]{\includegraphics[width=0.5\linewidth]{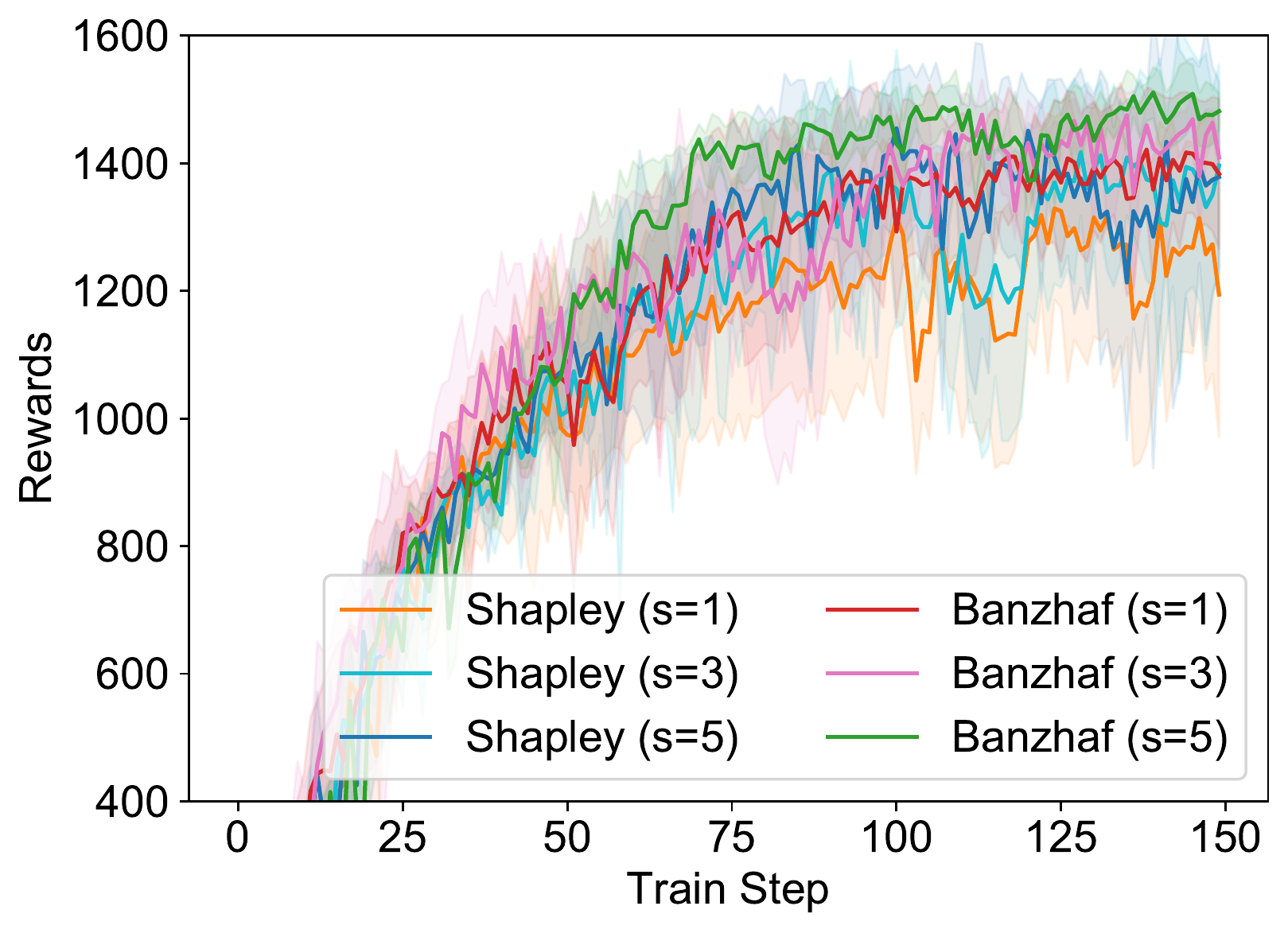}\label{fig:ant_sample_size}}%
	\caption{Results for Varying Sample Sizes}
\end{figure}

\subsubsection{Variation in Sample Sizes} Lastly, we examine how the number of coalitions sampled per agent per step (when estimating the semivalues) affects the robot's performance. We compare \mbshapley for the number of samples of $s=1,3,$ and $5$. We do the same for \mbbanzhaf. 
Figures~\ref{fig:cheetah_sample_size} and~\ref{fig:ant_sample_size} show that for both robots and both semivalue variants, reducing the number of sampled coalitions per node does not drastically decrease the performance in terms of reward and sample efficiency. 
Even using a small number of sampled coalitions, such as one per agent in each step, can still yield sufficiently good performance. We conjecture that this robustness property arises because, despite the small number of sampled coalitions per step, an RL algorithm typically runs on the scale of million timesteps. Hence, the number of sampled coalitions throughout the entire learning phase often will be sufficiently large. Importantly, because RL requires many credit assignment computations where each involves neural network forward propagation, this property helps us significantly reduce computation cost without compromising the learning performance.

\section{Related Work}


\noindent \textbf{Multiagent RL.} The simplest form of multiagent RL is Independent Q-learning (IQL)~\cite{tan1993multi}, where a group of agents each learns on their own and treats other agents as part of the environment. While IQL delivers decent performance, the agents frequently face the challenge that the environment appears unstationary, as other agents simultaneously learn and update their policies. To address this issue, centralised training, decentralised execution framework was first proposed separately by~\cite{foerster2018counterfactual} and ~\cite{lowe2017multi}. This training paradigm allows the agents to access the global information during training and has been a standard approach to recent multiagent RL algorithms since then. \cite{foerster2018counterfactual} proposed an actor-critic algorithm that addresses multiagent credit assignment explicitly using the counterfactual value of each agent and is applied to domains with discrete action space such as Starcraft. \cite{lowe2017multi} proposed a multiagent policy gradient algorithm where each agent receives a separate reward and uses their critic. This method can apply to competitive settings but does not address the multiagent credit assignment problem. Another line of work using centralised training, decentralised execution, and Q-learning are the implicit credit assignment methods such as~\cite{rashid2018qmix}~\cite{rashid2020weighted}~\cite{wang2020qplex}. More recently, \cite{yu2021surprising} showed the effectiveness of PPO in multiagent problems for discrete action space domains. Their model is optimised using a centralised reward and demonstrated to outperform baseline methods such as \cite{lowe2017multi}\cite{rashid2018qmix}.

\noindent \textbf{Game-theoretic Credit Assignment for Multiagent RL.} Several works have considered the Shapley value for credit assignment in multiagent RL with discrete action spaces~\cite{li2021shapley}\cite{wang2020shapley}\cite{wang2021shaq}\cite{xu2020learning}. Most notably among these works, \cite{li2021shapley} has shown state-of-the-art performance on Starcraft. In our experiments, we adapt their credit assignment framework from discrete domains to the continuous multiagent PPO. Unlike most prior credit assignment methods, which only consider one of the common values (e.g., Shapley value), we introduce semivalues to credit assignment in multiagent RL and define a more generic game-theoretic framework that encompasses a family of common solution concepts and analyzes the relation between them. Moreover, 
we use the new framework to address the problem of multiagent continuous control, and show that our model-based RL module can better estimate the coalition values for credit assignment and improve the sample efficiency accordingly.

\vspace{1ex}\noindent\textbf{Model-based RL.}
Model-based RL approaches typically alternate between fitting a predictive model of the environment dynamics/rewards and updating the control policies. The model can be used in various ways, such as execution-time planning~\cite{chua2018deep, nagabandi2018neural}, generating imaginary experiences for training the control policy~\cite{sutton1991dyna, janner2019trust}), etc. Our work is inspired by~\cite{feinberg2018model}, which addresses the problem of error in long-horizon model dynamics prediction. \cite{feinberg2018model} presents a hybrid algorithm that uses the model to simulate the short-term horizon and Q-learning to estimate the long-term value beyond the simulation horizon.
Unlike the previous model-based RL works, our work is the first to introduce model-based RL for enabling game-theoretic credit assignment in multiagent continuous control.

\section{Conclusions}
In this paper, we studied multiagent robotic continuous control with model-based game-theoretic credit assignments. We first modelled robot joints as a fully-cooperative multiagent system, and optimised the system using a multiagent version of PPO. We then proposed a generic game-theoretic credit assignment framework using semivalues for evaluating agent-specific advantage functions. Furthermore, we proposed a model-based framework which significantly improved estimation of coalition values, which empowers game-theoretic credit assignments for multiagent continuous control. Finally, we empirically demonstrate that our model-based credit assignments leads to sample-efficient and robust multiagent learning on MuJoCo robot locomotion tasks.

In the future, we would like to study other continuous control algorithms and other robot variants, including real robots. Furthermore, it would interesting to study the model-based RL methods with other configurations such as longer prediction horizons.



\bibliographystyle{plain}  
\bibliography{sample}


\end{document}